\documentclass[letterpaper]{article} 
\usepackage{aaai25}  
\usepackage{times}  
\usepackage{helvet}  
\usepackage{courier}  
\usepackage[hyphens]{url}  
\usepackage{graphicx} 
\usepackage{natbib}  
\usepackage{caption} 
\usepackage{algorithm}

\usepackage{newfloat}
\usepackage{listings}
\DeclareCaptionStyle{ruled}{labelfont=normalfont,labelsep=colon,strut=off} 
\floatstyle{ruled}
\newfloat{listing}{tb}{lst}{}
\floatname{listing}{Listing}
\usepackage{tabularx}
\usepackage{array}
\usepackage{amsmath}
\usepackage{amssymb}
\usepackage{amsthm}
\usepackage{algorithm}
\usepackage{algpseudocode}
\usepackage{verbatim}
\usepackage{multirow}

\newcommand{\nf}[1]{\|#1\|_{F}}
\newcommand{\Rnm}{\mathbb R^{n\times m}}
\newcommand{\Ib}{\boldsymbol{I}}
\newcommand{\nz}[1]{\|#1\|_{0}}
\newcommand{\trainmat}{\Xb~=~[\xb_1,\xb_2, \ldots\xb_{m}] \in \mathbb{R}^{n\times m}}
\newcommand{\Xb}{\boldsymbol{X}}
\newcommand{\Eb}{\mathbb{E}}
\newcommand{\Sigmab}{\boldsymbol{\Sigma}}
\newcommand{\sigmab}{\boldsymbol{\sigma}}
\newcommand{\Db}{\boldsymbol{D}}
\newcommand{\db}{\boldsymbol{d}}
\newcommand{\Rn}{\mathbb R^n}
\newcommand{\Rnn}{\mathbb R^{n\times n}}
\newcommand{\xnset}{\{x_1, x_2, \ldots x_n\}}

\newcommand{\ynset}{\{y_1, y_2, \ldots y_n\}}
\newcommand{\xnvec}{\boldsymbol{x} = [x_1, x_2, \ldots, x_n]^T}
\newcommand{\xb}{\boldsymbol{x}}

\newcommand{\thup}[1]{#1^\textrm{th}}

\newcommand{\PA}{\mathcal{P}}
\newcommand{\oneton}{\in\{1,2,\ldots,n\}}
\newcommand\independent{\protect\mathpalette{\protect\independenT}{\perp}}
\def\independenT#1#2{\mathrel{\rlap{$#1#2$}\mkern2mu{#1#2}}}
\newcommand{\Wb}{\boldsymbol W}
\DeclareMathOperator*{\argmax}{argmax}
\DeclareMathOperator*{\argmin}{argmin}
\theoremstyle{definition}
\newtheorem{exmp}{Example}
\newtheorem{thm}{Theorem}

\algnewcommand{\Inputs}[1]{%
	\State \textbf{Inputs:}
	\Statex \hspace*{\algorithmicindent}\parbox[t]{.8\linewidth}{\raggedright #1}
}
\algnewcommand{\Initialize}[1]{%
	\State \textbf{Initialize:}
	\Statex \hspace*{\algorithmicindent}\parbox[t]{.8\linewidth}{\raggedright #1}
}

\algnewcommand{\Calculations}{%
	\State \textbf{Calculations:}
	\Statex 
}

\title{Induced Covariance for Causal Discovery in Linear Sparse Structures}
\author{
	Saeed Mohseni-Sehdeh\textsuperscript{\rm 1}, 
	Walid Saad\textsuperscript{\rm 1}, 
}
\affiliations {
	\textsuperscript{\rm 1}Wireless@ VT, Bradly Department of Electrical and Computer Engineering, Virginia Tech, Arlington, VA, USA\\
	saeedmohseni@vt.edu, walids@vt.edu
}

\begin{document}

	\date{}
	\maketitle
	
	\begin{abstract}
		
		Causal models seek to unravel the cause-effect relationships among variables from observed data, as opposed to mere mappings among them, as traditional regression models do. This paper introduces a novel causal discovery algorithm designed for settings in which variables exhibit linearly sparse relationships.
		In such scenarios, the causal links represented by directed acyclic graphs (DAGs) can be encapsulated in a structural matrix. The proposed approach leverages the structural matrix's ability to reconstruct data and the statistical properties it imposes on the data to identify the correct structural matrix. 
		This method does not rely on independence tests or graph fitting procedures, making it suitable for scenarios with limited training data. Simulation results demonstrate that the proposed method outperforms the well-known PC, GES, BIC exact search, and LINGAM-based methods in recovering linearly sparse causal structures.
		
	\end{abstract}

	\section{Introduction}\label{sec.introduction}
	Causal learning is an approach used to extract and understand cause-and-effect relationships from data. This approach seeks to uncover the fundamental structures that determine how data are related \cite{causal_relation_interest}. This structural understanding is at a deeper level than that observed in statistical learning, which is focused on learning various mappings among data \cite{stat2cause}.
	Discovering causal relations plays a crucial role in the scientific method \cite{camps2023discovering}. A comprehensive causal model of a phenomenon could describe the observed data and consistently make predictions. The advantage of this type of learning over statistical learning, which identifies mere associations between variables, lies in its generalization and robustness to distribution changes \cite{towardcausrep}. Furthermore, causal relations may be transferable to other problems, which constitutes an additional benefit \cite{towardcausrep}. The main challenge, however, lies in effectively discovering causal relationships form observed data.
	
	Several approaches have been proposed for causal discovery \cite{IC, SGS, lingamIC, lingamDirect, meek, chickering, exactsearch}. These methods are generally classified into two categories \cite{stat2cause}: constraint-based and score-based methods. In constraint-based methods, conditional independencies among variables are tested and the inferred relations are represented using a DAG that best reflects them. Notable examples of such algorithms include inductive causation (IC) \cite{IC}, Spirtes-Glymour-Scheines (SGS) \cite{SGS}, Peter-Clark (PC) \cite{SGS} and linear non-Gaussian acyclic model (LINGAM) based methods \cite{lingamIC} and \cite{lingamDirect}. IC and SGS algorithms examine the conditinal independencies between each pair of variables conditioned on any subset of the remaining variables and use this information for forming the causal graph. This approach can be computationally intensive due to the large number of subsets. The PC algorithm mitigates this challenge by initiating the search from a complete graph and sequentially removing edges systematically testing the conditinal independencies of each pair and their naighbours. Although the PC algorithm reduces computational cost, it still depends on conditional independence tests, which are computationally demanding and require substantial data, particularly in high-dimensional settings, to produce reliable results.
	Unlike the previously mentioned methods that rely heavily on conditional independence tests, LiNGAM-based approaches assume non-Gaussianity and linearity in the data. By leveraging these assumptions, LiNGAM aims to identify the causal graph. Although alleviating the challenge of conditional tests, the non-Gaussianity is a limitting assumption.
	
	Alternatively, score-based methods use a scoring function to evaluate graphical representations. Possible graphs are tested against the data, and the graph with the highest score is selected. Some of the prominent methods in this category are greedy equivalent search (GES) \cite{meek, chickering}, and Bayesian information criterion (BIC) exact search \cite{exactsearch}. The primary drawback of these methods \cite{meek, chickering, exactsearch} is the exponential growth of the possible graphs as the number of variables (nodes of the graph) increase, which result in higher computation demands. Hence, clearly, despite the emergence of several approaches for causal discovery, there remains a lack of a scalable algorithm that is well-suited for scenarios involving small datasets.
	
	The main contribution of this paper is a novel causal discovery algorithm tailored for linear sparse structures. Linear structures are known for their interpretability and scalability. They are widely utilized in applications such as time series analysis \cite{TS}. We introduce an algorithm that recovers the structural matrix, which encapsulates causal graph information, by exploiting the induced covariance, data recovery capability, rank, and diagonal structure of the matrix for scenarios in which the relationships between variables can be effectively modeled by linear sparse dependencies. Unlike traditional methods \cite{IC, SGS, lingamIC, lingamDirect, meek, chickering, exactsearch}, the proposed algorithm does not rely on independence tests or graph fitting procedures, making it less dependent on large datasets. Our simulations demonstrate its effectiveness in various germane scenarios. The results show that our algorithm outperforms the aforementioned methods by an average of 35\% in precision and 41\% in recall.

	\section{The Proposed Algorithm}\label{sec.method}
	For stating the problem, we first review the concept of a structural causal model (SCM), a popular method for causal relations modelling \cite{stat2cause}. In this framework, a set of random variables $\ynset$ is represented as the vertices of a directed acyclic graph (DAG), and the following relations hold:
	\begin{equation}
		y_i = f_i(\PA_i, u_i) \quad \forall i \oneton,
	\end{equation} 
	with $f_i$ being a deterministic function, $\PA_i$ (parents) represents the set of variables that influence $y_i$, and $u_i$ is an unexplained noise random variable \cite{towardcausrep}. In the graphical SCM representation, a directed edge exists from each member of $\PA_i$ to $y_i$ for $i \oneton$. The process of causal discovery involves identifying $f_i$ and $\PA_i$ for all $n \oneton$.

	\subsection{Problem statement}
	Let $\xnvec \in\Rn$ be a vector containing all the random variables for which the causal relationships are to be discovered. We define $\mathcal{I}$ as the set of indices for independent variables and $\mathcal{D}$ as the set of indices for dependent variables.
	According to SCM model, each $x_i$ for $i \in \mathcal{D}$ is a function of a subset of independent variables $\PA_i$, which are considered the parents of $x_i$. We assume that functions $f_i, \forall i \oneton$
	are linear and $|\PA_i| \leq \tau$, where $|.|$ represents the cardinality of a set and $\tau$ is a model parameter. Under these assumptions, we can represent $x_i$ as follows:
	\begin{equation}
		x_i = \db_i^T \xb,
	\end{equation}
	where $\db_i \in \Rn$ is a vector with fewer than $\tau$
	 non-zero elements, corresponding to the independent variables upon which $x_i$ depends. Given this notation, the relationship for all variables can be expressed as
	\begin{equation} \label{eq.1}
		\xb = \Db \xb,
	\end{equation}
	where $\Db \in \Rnn$ is a matrix constructed from the vectors $\db_i, \forall i\oneton$ as its rows. Matrix $\Db$ contains all pertinent information regarding the causal structure of this model, and an accurate estimation of $\Db$ reveals the underlying causal relations.
	
	In practice, there is often limited or no prior knowledge about the underlying structure of the data, and only the dataset itself is available. We use $\trainmat$ to represent the given dataset, where each $\xb_i \in \Rn$ is a sample. By applying (\ref{eq.1}), we have
	\begin{equation}
		\Xb = \Db \Xb.
	\end{equation}	
	The primary objective is to determine the causal structure ($\Db$) from $\Xb$. Estimation of $\Db$ does not involve conditional independence tests, making it suitable when the number of data samples are limited, especially in high dimensional data.
	\subsection{Challenges}
	In the estimation of $\Db$, several challenges must be addressed. Based on the previous discussion, it can be inferred that $\Db$ must satisfy the condition expressed in (\ref{eq.1}). However, as demonstrated in the following example, this condition alone is insufficient for uniquely determining the causal structure.
	\begin{exmp}
		Suppose data is created as follows
		\begin{equation}
			\begin{bmatrix}
				x_1 \\
				x_2 \\
				x_3
			\end{bmatrix}
			=
			\begin{bmatrix}
				1 & 0 & 0 \\
				0 & 1 & 0 \\
				1 & 1 & 0
			\end{bmatrix}
			\begin{bmatrix}
				x_1 \\
				x_2 \\
				x_3
			\end{bmatrix}.
		\end{equation}
		This structure suggests that $x_1$ and $x_2$
		are independent variables (as they are not linear combinations of any other variables), while $x_3$ is the sum of $x_1$ and 
		$x_2$. When only the data is available and the objective is to satisfy the condition given in (\ref{eq.1}), the solution may not be unique. For example, the identity matrix ($\Ib \in \mathbb{R}^{3\times 3}$) and the following matrix also satisfies (\ref{eq.1}):

		\begin{equation}
			\begin{bmatrix}
				0 & -1 & 1 \\
				0 & 1 & 0 \\
				0 & 0 & 1
			\end{bmatrix}.
		\end{equation}
		This is a common problem in causal discovery as multiple graphs can describe the same data \cite{SGS}. This example also illustrates how causal discovery differs from a regression problem. While both solutions may be acceptable in the context of regression, only one solution reveals the underlying causal structure. In other words, what separates this approach from a linear algebra regression is that we are not looking for any solution of (\ref{eq.1}) but the one that describes the associations according to cause-effect relations. 
	\end{exmp}
	
	\subsection{Solutions}
	To address the challenges discussed in the previous subsection, it is beneficial to explore certain properties related to the matrix
	$\Db$, which represents the causal relations.

	The structure of $\Db$ can provide valuable insights into the relations among variables. Any $\thup{k}$ row of the structural matrix $\Db$,  whose all elements are all equal to zero except for the element in column $k$, which is equal to $1$, corresponds to an independent random variable.
	Since all variables are linear combinations of these  variables, the rank of $\Db$ corresponds to the number of independent variables. This observation suggests that the structure of $\Db$ can be used to infer the statistical relationships between the variables. This property is particularly useful for narrowing down the number of potential solutions for $\Db$.
	
	To further constrain the possible solutions for $\Db$, the following theorem establishes a connection between the statistical properties of the data and the structural matrix. More specifically it shows that selcting a specific value for variable $\Db$, uniquely determines the value of covariance matrix of data, indicating that $\Db$ imposes a constraint on covariance matrix resulting in \emph{induced covariance}.
	\begin{thm}
		Consider $\Db \in \Rnn$ to be a matrix that represents a linear causal structure governing the zero mean variables $\xnvec \in \Rn$. The covariance matrix of these variables is given by $\Db\sigmab\Db^T$, in which $\sigmab \in \Rnn$ is a diagonal matrix with diagonal elements being the variance of variables $\xb$.
		\begin{proof}
			Let $\Eb$ be the expected value operator.
			To prove this theorem, we derive an expression for
			$\Eb[x_i,x_j]$ where $x_i$ and $x_j$ are two components of the random vector $\xb$. Based on the causal structure, following holds
			\begin{align}
				x_i = \db_i^T\xb,\\
				x_j = \db_j^T\xb,
			\end{align}
			in which $\db_i^T$ and $\db_j^T$ are the rows $i$ and $j$ of the structural matrix $\Db$. Thus,
			\begin{equation}
				\Eb[x_i,x_j] = \Eb[\db_i^T\xb\db_j^T\xb].
			\end{equation}
			Since the only nonzero elements in  $\db_i^T\xb\db_j^T\xb$ occurs when both $\db_j$ and $\db_i$ have non-zero elements in the same positions, then
			\begin{align}
				\Eb[x_i,x_j] = \db_i^T\sigmab\db_j^T.
			\end{align}
			By applying the same procedure to all $(i,j)$ pairs, the theorem is proven.
		\end{proof} 
	\end{thm}
	This theorem restricts the solutions to (\ref{eq.1}) by imposing that the correct solution must not only satisfy (\ref{eq.1}) but also fulfill the condition $\Sigmab = \Db\sigmab\Db^T$, 
	where $\Db\sigmab\Db^T$ is the induced covariance by $\Db$ and 
	$\Sigmab \in \Rnn$ is the original covariance matrix of data, which can be estimated directly from data.
	
	By using the properties of $\Db$ and its implications on the structure of data, we can formulate the following optimizatiopn problem for structure recovery:
	\begin{equation}
		\begin{aligned}
			\argmin_{\Db}  \{\textrm{rank($\Db$)} &+ \lambda \textrm{Tr($\Db$)\} }\\
			\text{subject to} \quad & \Xb = \Db\Xb, \\
			& \Sigmab = \Db\sigmab\Db^T, \\
			& \nz{\db_i^T}\leq\tau \quad \forall i \oneton. \\
		\end{aligned}
	\end{equation}
	In this formulation, $\Db \in \mathbb{R}^{n \times n}$ represents the structural matrix, while $\trainmat$ is the dataset. The covariance matrix of the data is represented by $\Sigmab \in \mathbb{R}^{n \times n}$, and $\sigmab \in \mathbb{R}^{n \times n}$ is a diagonal matrix whose diagonal entries correspond to those of $\Sigmab$. The term $\db_i^T$ represent row $i$ of $\Db$. The operator $\nz{\cdot}$ returns the number of non-zero elements in a vector. Additionally, $\tau$ is the maximum allowable number of non-zero elements in each row, and $\lambda$ serves as a scaling parameter. The rank($\Db$) term prevents the model from overcomplication and Tr($\Db$) prevents the the trivial $\Ib$ solution.
	
	Rank (the number of non-zero singular values) requires combinatory calculation which makes the problem untractable. To address this challenge, the idea proposed in \cite{l0} is used, which approximate $\nz{.}$ as:
	\begin{equation}
		\nz{x} \approx 1 - e^{-\frac{x^2}{\sigma^2}}.
	\end{equation}
	
	By combining these ideas, the final problem formulation is
	\begin{equation}\label{eq.2}
		\begin{aligned}
			\argmin_D  \{\sum_{i=1}^{n}(1-e^{-\frac{s_i^2}{\sigma^2}}) &+ \lambda \sum_{i=1}^{n}(1-e^{-\frac{d_{(i,i)}^2}{\sigma^2}})\} \\
			\text{subject to} \quad & \Xb = \Db\Xb, \\
			& \Sigmab = \Db\sigmab\Db^T, \\
			& \nz{\db_i^T}\leq\tau \quad \forall i \oneton,\\
		\end{aligned}
	\end{equation}
	where $s_i, \forall i\oneton$ are the singular values of $\Db$.
	To present the final algorithm for obtaining the solution of (\ref{eq.2}), it is necessary to consider $ \nz{\db_i^T}\leq\tau \quad \forall i \in \{1,\ldots,n\}$, which also requires combinatory calculations. To  handle that, we propose solving the following optimization problem:
	
	\begin{equation} \label{eq.3}
		\begin{aligned}
			\argmin_D  \{\sum_{i=1}^{n}(1-e^{-\frac{s_i^2}{\sigma^2}}) &+ \lambda \sum_{i=1}^{n}(1-e^{-\frac{d_{(i,i)}^2}{\sigma^2}})\} \\
			\text{subject to} \quad & \Xb = \Db\Xb, \\
			& \Sigmab = \Db\sigmab\Db^T, \\
		\end{aligned}
	\end{equation}
	and for each row of the resulting $\Db$, we retain only the $\tau$ entries with the largest absolute values. This process is iterated $N$ times. Due to noise effects on data, (\ref{eq.3}) might not have a solution, therefore some relaxation on constraint might be required. This is done as follows:
	\begin{equation} 
		\begin{aligned}
			\argmin_D  \{\sum_{i=1}^{n}(1-e^{-\frac{s_i^2}{\sigma^2}}) &+ \lambda \sum_{i=1}^{n}(1-e^{-\frac{d_{(i,i)}^2}{\sigma^2}})\} \\
			\text{subject to} \quad & \nf{\Xb - \Db\Xb}^2\leq\epsilon_1, \\
			& \nf{\Sigmab - \Db\sigmab\Db^T}^2\leq\epsilon_2, \\
		\end{aligned}
	\end{equation}
	where $\epsilon_1$ and $\epsilon_2$ can be tuned to result in the best result.
	
	Solving (\ref{eq.2}) requires an initial estimate for $\Db$, and the final value of the objective function is dependent on this initial estimate. To obtain the optimal solution, we propose executing the algorithm multiple times, each with a distinct random initial estimate. The solution yielding the lowest value of the objective function is then retained as the final result.

	The complete sparse linear causal discovery (SLCD) algorithm is presented in Algorithm \ref{alg.1}.
	
	\section{Simulation Results and Analysis }\label{sec.simulation}
	This section presents the results of our simulation studies. We compare our method with PC, GES, LINGAM IC, LINGAM Direct and BIC exact search for performance evaluation.
	For comprehensive reporting, we evaluate following metrics: the data reconstruction error, the recovery error of the causal matrix, the recovery error of the covariance matrix, precision (porportion of the number of correct estimated links to the total number of estimated links), and recall (porportion of the number of correct estimated links to the total number of links in the true graph). Let $\hat{\Db} \in \mathbb{R}^{n \times n}$ be the estimated matrix obtained from the proposed algorithm, and $\Xb \in \mathbb{R}^{n \times m}$ represent the training data. The reconstruction error is then defined as:
	\begin{equation}
		\frac{1}{nm}\nf{\Xb - \hat{\Db}\Xb}^2.
	\end{equation}
	
	We define the true structural matrix as $\Db \in \Rnn$ and, thus, the recovery error of structural matrix will be:
	\begin{equation}
		\frac{1}{n^2}\nf{\Db - \hat{\Db}}.
	\end{equation}
	
	Let $\Sigmab \in \Rnn$ represent the true covariance matrix of the original data with $\Sigmab \in \Rnn$, then the recovery error of the covariance matrix is defined as
	\begin{equation}
		\frac{1}{n^2} \nf{\Sigmab - \hat{\Db}\sigma\hat{\Db}^T},
	\end{equation}
	in which, $\sigmab$ is a diagonal matrix with diagonal elements of $\Sigmab$. 
	
		\begin{table*}
		\centering
		\begin{tabular}{|c|c|c|c|c|c|c|c|c|}
			\hline
			\textbf{Dataset} & \textbf{IV count}\footnote{Independent variable count} & \textbf{$x_1$} & \textbf{$x_2$} & \textbf{$x_3$} & \textbf{$x_4$} & \textbf{$x_5$} & \textbf{$x_6$} &
			\textbf{$x_7$}
			\\
			\hline
			Dataset 1 & 1 & U(-2.5, 2.5) &  $2x_1$ & $0.4x_1$ & -& -& -&-\\
			\hline
			Dataset 2 & 2 & U(-2.5, 2.5) & U(-2.5, 2.5) & $0.3x_1$ & $x_1 + 2x_2$&- &-&- \\
			\hline
			Dataset 3 & 2 & U(-2.5, 2.5) & U(-2.5, 2.5) & $x_1 + 3x_2$ & $2x_2$ & $2x_1 + x_2$ & -&-\\
			\hline
			Dataset 4 & 3 & U(-2.5, 2.5)  & U(-2.5, 2.5)  & N(0,4) & $x_1 + 0.3x_3$& $2x_1 + 3x_2$ & $2x_2 + 0.5x_3$&- \\
			\hline
			
			Dataset 5 & 3 & U(-2.5, 2.5)  & U(-2.5, 2.5)  & N(0,4) & $x_1 + 0.5x_3$& $x_2 + 2x_3$ & $x_1 + 3x_3$&$x_2+x_3$ \\
			\hline
		\end{tabular}
		\caption{Datasets Information.}
		\label{tab.1}
	\end{table*}
	
	For simulation purposes, five distinct datasets were generated, henceforth referred to as Dataset 1, Dataset 2, Dataset 3, and Dataset 4, Dataset 5. Each dataset comprises 1000 samples. Table \ref{tab.1} provides detailed information on the generation process for each dataset. 
	The variables $x_i$, where $i \in \{1,2,\ldots,6\}$, represent the elements of the data vector $\xb = [x_1, x_2, \ldots, x_6]^T$. The presence of a '-' symbol in place of a variable indicates its absence from the corresponding dataset, reflecting the varying dimensionality across datasets. The table indicates the data distribution from which the samples of independent variables are drawn. For dependent variables, the table specifies the linear combinations used to generate them.
	
	 $U(a,b)$ represents the uniform distribution of data in the $[a,b]$ interval. $N(\mu, \sigma^2)$ represents a Gaussian random variable with mean $\mu$ and variance $\sigma^2$.
	By constructing the datasets in this manner, the variables exhibit the linear sparse relations that SLCD is specifically designed to handle. This approach also enables the evaluation of algorithm performance across various data dimensions. Additionally, the datasets includes independent variables with different data distributions, allowing for the assessment of algorithm robustness under diverse distributional scenarios. It is important to note that for the dependent variables, each linear combination results in a convolution of the data distributions, further contributing to the variability in the distributions of the dataset's variables.
	
\begin{figure}
		\begin{center}
			\includegraphics[scale=.5]{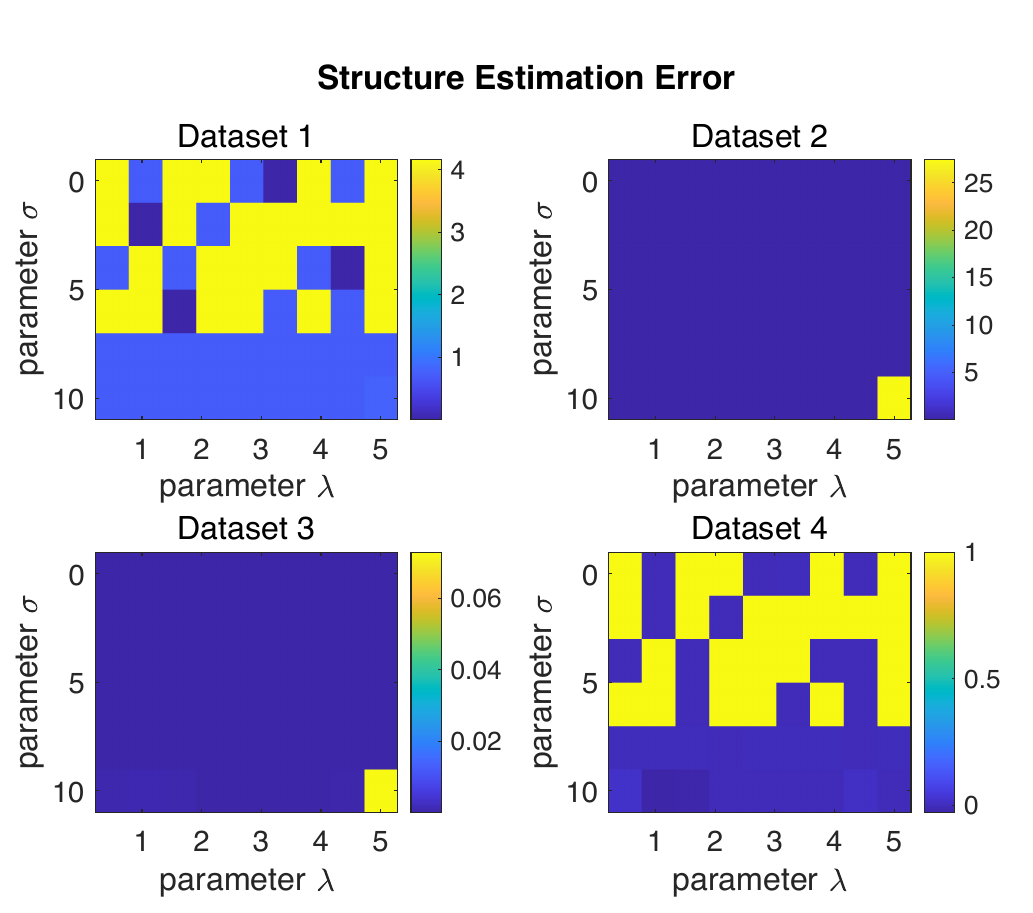}
			\caption{Structure estimation error for various datasets and various hyperparameters.}
			\label{fig.1}
		\end{center}		
	\end{figure}

\begin{figure}
		\begin{center}
			\includegraphics[scale=.5]{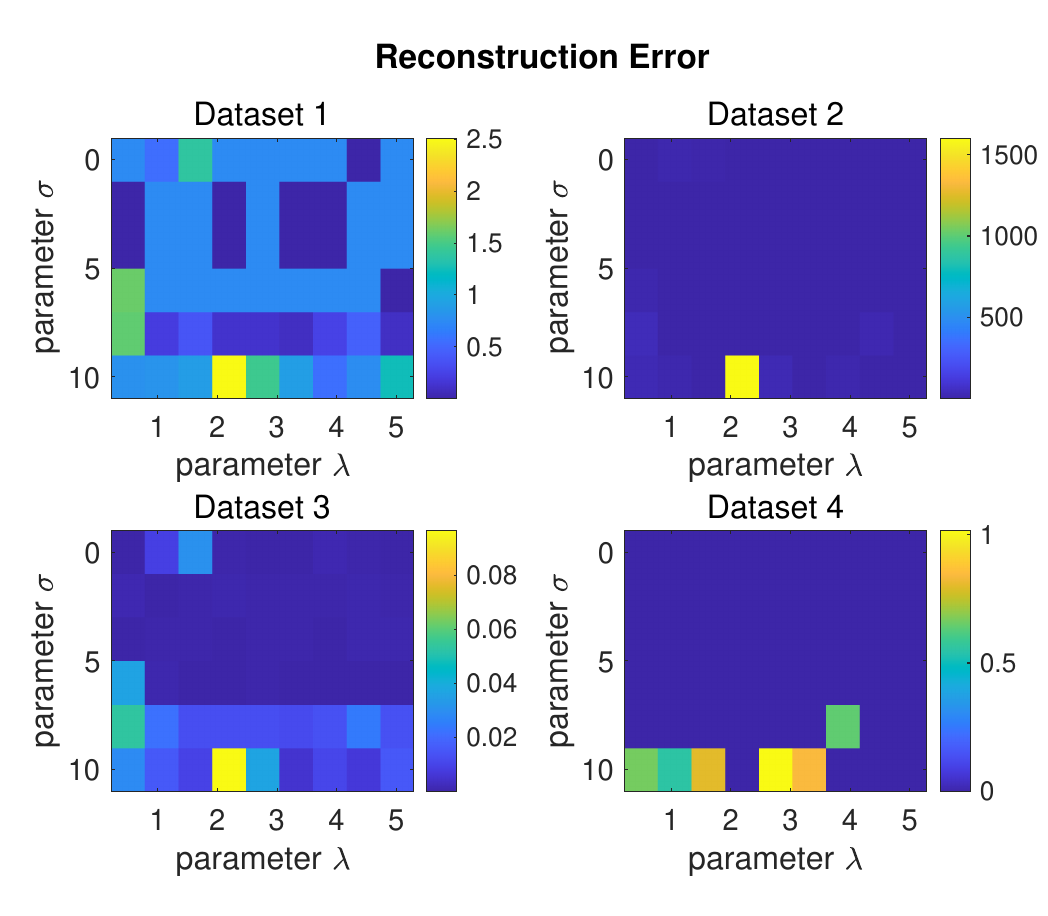}
			\caption{Reconstruction error for various datasets and various hyperparameters.}
			\label{fig.2}
		\end{center}
	\end{figure}

\begin{figure}
		\begin{center}
			\includegraphics[scale=.5]{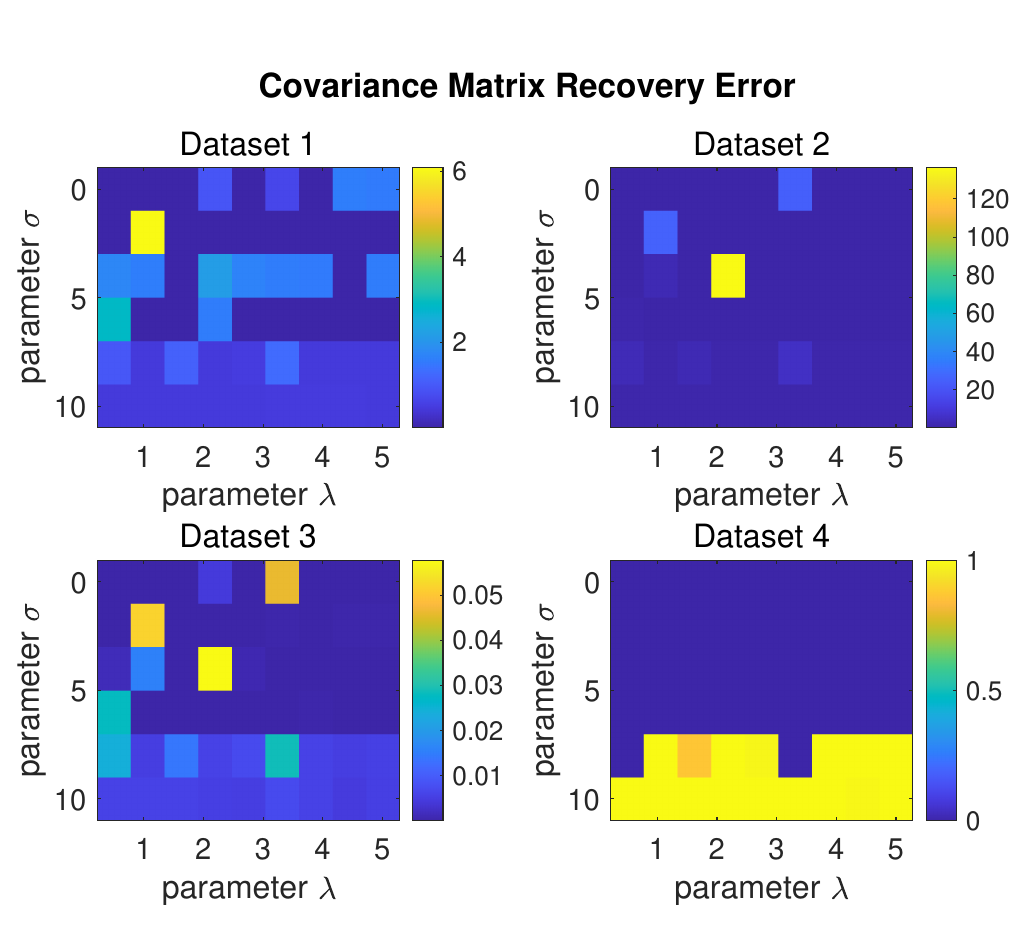}
			\caption{Covariance matrix estimation error for various datasets and various hyperparameters.}
			\label{fig.3}
		\end{center}
	\end{figure}

	Figures \ref{fig.1} through \ref{fig.3} display the algorithm's simulation results, highlighting how different hyperparameter settings influence performance metrics. The table shows metrics for a specific hyperparameter set. The results reveal moderate sensitivity to hyperparameters, with effective recovery of the underlying structure when parameters are chosen appropriately. The figures also indicate a broad range of satisfactory parameters, demonstrating the method's robustness.
	
	Table \ref{tab.2} shows the output of SLCD method for each dataset with $(\sigma, \lambda) = (0.3,5)$. It shows that the method successfully recovers the structural matrix, with the exception of Dataset 1. 
	
	Table \ref{tab.3} presents the simulation results of SLCD in comparison with several well-known causal discovery algorithms. The results indicate that SLCD outperforms the other methods by an average of $35\%$ in precision and $41.5\%$ in recall across Datasets 2 through 5. While all methods exhibit challenges with Dataset 1, SLCD consistently demonstrates superior performance in the remaining datasets.
	
	 The method's performance is suboptimal in this scenario. This can be attributed to the structure of Dataset 1, wherein only one independent variable exists, and all other variables are scalar multiples thereof. This configuration does not provide sufficient information to unambiguously identify the independent variable, as any of the variables could potentially fulfill this role. This ambiguity introduces uncertainty into the algorithm, potentially leading to diverse solutions. However, as the structural complexity increases with the introduction of additional independent variables, the informational content of the data becomes more robust, facilitating more accurate recovery of the underlying causal structure.

	\begin{algorithm}[t]
		\caption{Sparse Linear Causal Discovery (
			SLCD) Algorithm.}\label{alg.1}
		\begin{algorithmic}
			\Inputs{$\Xb \in \Rnm$, $N$, $M$, $\lambda$, $\sigma$, $\tau$} 
			\For{$t = 1:M$}
			\Initialize{$\Db_0 \in \Rnn: randomly$}
			\If{(t == 1)}
			\State$J_{min} \gets J(\Db_0)$
			\State$\Db_{opt} \gets \Db_0$
			\EndIf
			\For{$k = 1:N$}
			\State $\Db \gets$ Solve (\ref{eq.3}) (e.g. fmincon (MATLAB))
			\If{$J_{min} > J(\Db)$}
			\State$J_{min} \gets J(\Db)$
			\State$\Db_{opt} \gets \Db$
			\EndIf
			\EndFor
			\EndFor\\
			\Return $\Db_{opt}$
		\end{algorithmic}
	\end{algorithm}    
	
	\section{Conclusion}\label{sec.conclusion}
	This paper proposes an algorithm for causal discovery within a linear sparse structure. The algorithm leverages properties induced by the causal structure matrix on the data to facilitate its recovery. Specifically, it utilizes the rank of the matrix, which indicates the number of independent variables, and the concept of induced covariance.
	Simulation studies demonstrate the algorithm's efficacy across various configurations. However, there are two potential avenues for future research:
	\begin{enumerate}
		\item Generalization to non-linear setups: Extending the algorithm's applicability beyond linear causal relationships could significantly broaden its utility in real-world scenarios where complex, non-linear interactions are prevalent.
		\item Refinement of independence criteria: While the induced covariance proved helpful in determining variable independence, it does not guarantee a definitive determination of independence. Future work could focus on developing more robust constraints for this purpose, potentially incorporating information-theoretic measures or advanced statistical techniques to more accurately characterize independence relationships.
	\end{enumerate}
	These extensions could enhance the algorithm's versatility and reliability in uncovering causal structures across a wider range of applications and data types.

	\begin{table*}
		\centering
		\begin{tabular}{|c|c|c|c|c|c|}
			\hline
			\textbf{Dataset} & \textbf{Structure matrix} & \textbf{Estimated structure matrix} & \textbf{($\sigma, \lambda$)} \\
			\hline
			Dataset 1 & $\begin{bmatrix}
				1 & 0 & 0 \\
				2 & 1 & 0 \\
				0.4 & 1 & 0
			\end{bmatrix}$  & $\begin{bmatrix}
				0 & 0.5 & 4.4\times10^{-7} \\
				2 & 1 & 1.1\times10^{-6} \\
				-1.2\times10^{-7} & 0.2 & 0
			\end{bmatrix}$ & (0.3, 5) \\
			\hline
			Dataset 2 & $\begin{bmatrix}
				1 & 0 & 0 & 0 \\
				0 & 1 & 0 & 0\\
				0.3 & 1 & 0 & 0\\
				1 & 2 & 0 & 0\\
			\end{bmatrix}$ & $\begin{bmatrix}
				1 & 0 & -8.8\times10^{-4} & 0 \\
				0 & 1 & -3.3\times10^{-4} & 0\\
				0.3 & 0 & -2.7\times10^{-4} & 0\\
				1.002 & 1.999 & 0 & 0\\
			\end{bmatrix}$ & (0.3, 5) \\
			\hline
			Dataset 3 & $\begin{bmatrix}
				1 & 0 & 0 & 0 &0\\
				0 & 1 & 0 & 0&0\\
				1 & 3 & 0 & 0&0\\
				0 & 2 & 0 & 0 &0\\
				2 & 1 & 0 & 0 & 0
			\end{bmatrix}$ & $\begin{bmatrix}
				0.999 & 0.0497 & 0 & 0 &0\\
				0 & 1.000 & 0 & 0&0.0102\\
				0.976 & 3.049 & 0 & 0&0\\
				-0.0147 & 1.999 & 0 & 0 &0\\
				1.990 & 1.099 & 0 & 0 & 0
			\end{bmatrix}$ & (0.3, 5) \\
			\hline
			Dataset 4 & $\begin{bmatrix}
				1 & 0 & 0 & 0 &0&0\\
				0 & 1 & 0 & 0&0&0\\
				0 & 0 & 1 & 0&0&0\\
				1 & 0 & 0.3 & 0 &0&0\\
				2 & 3 & 0 & 0 & 0&0\\
				0 & 2 & 0.5&0 &0&0
			\end{bmatrix}$ & $\begin{bmatrix}
				0.999 & -0.009 & 0 & 0 &0&0\\
				0.016 & 0.999 & 0 & 0&0&0\\
				-.0432 & 0 & 0.997 & 0&0&0\\
				0.987 & 0 & 0.3019 & 0 &0&0\\
				2.048 & 2.982 & 0 & 0 & 0&0\\
				0 & 1.995 & 0.483&0 &0&0
			\end{bmatrix}$ & (0.3, 5) \\
			\hline
			Dataset 5 & $\begin{bmatrix}
				1 & 0 & 0 & 0 &0&0&0\\
				0 & 1 & 0 & 0&0&0&0\\
				0 & 0 & 1 & 0&0&0&0\\
				1 & 0 & 0.5 & 0 &0&0&0\\
				0 & 1 & 2 & 0 & 0&0&0\\
				1 & 0 & 3&0 &0&0&0\\
				0&1&1&0&0&0&0
			\end{bmatrix}$ & $\begin{bmatrix}
				0.997&.0525&0&0&0&0&\\
				-0.082 & 0.994 & 0 & 0&0&0&0\\
				0.057 & 0 & 0.998 & 0&0&0&0\\
				1.025 & 0 & 0.491 & 0 &0&0&0\\
				0 & 0.956 & 2.024 & 0 & 0&0&0\\
				1.168 & 0 & 2.986&0 &0&0&0\\
				0&0.975&1.025&0&0&0&0
			\end{bmatrix}$ & (0.3, 5) \\
			\hline
		\end{tabular}
		\caption{True structural matrix and the output of SLCD.}
		\label{tab.2}
	\end{table*}

	\begin{table*}
		\centering
		\begin{tabular}{|c|c|c|c|c|c|c|}
			\hline
			\# & PC & GES & LINGAM IC & LINGAM Direct & BIC Search & SLCD \\
			\hline
			Precision   & 0.33   & 0.5   & 0   & 0.33   & 0   & 0    \\
			\hline
			Recall   & 1  & 0.5  & 0  & 0.5  & 0  & 0  \\
			\hline
			Number of Correct link estimation  & 2  & 1  & 0  & 1  & 0  & 0   \\
			\hline
			\multicolumn{7}{|c|}{Dataset 1}\\
			\hline
			\# & PC & GES & LINGAM IC & LINGAM Direct & BIC Search & SLCD \\
			\hline
			Precision   & 0.5   & 0.6   & 0.25   & 0   & 0.75   & 1    \\
			\hline
			Recall   & 0.66  & 1  & 0.33  & 0  & 1  & 1  \\
			\hline
			Number of Correct link estimation  & 2  & 3  & 1  & 0  & 3  & 3   \\
			\hline
			\multicolumn{7}{|c|}{Dataset 2}\\
			\hline
			\# & PC & GES & LINGAM IC & LINGAM Direct & BIC Search & SLCD \\
			\hline
			Precision   & 0.37   & 0.43   & 0   & 0   & 0.43   & 1    \\
			\hline
			Recall   & 0.6  & 0.6  & 0  & 0  & 0.6  & 1  \\
			\hline
			Number of Correct link estimation  & 3  & 3  & 0  & 0  & 3  & 5   \\
			\hline
			\multicolumn{7}{|c|}{Dataset 3}\\
			\hline
			\# & PC & GES & LINGAM IC & LINGAM Direct & BIC Search & SLCD \\
			\hline
			Precision   & 1   & 1   & 0.2   & 0.1   & 0.67   & 1    \\
			\hline
			Recall   & 1  & 1  & 0.33  & 0.17  & 1  & 1  \\
			\hline
			Number of Correct link estimation  & 6  & 6  & 2  & 1  & 6  & 6   \\
			\hline
			\multicolumn{7}{|c|}{Dataset 4}\\
			\hline
			\# & PC & GES & LINGAM IC & LINGAM Direct & BIC Search & SLCD \\
			\hline
			Precision   & 0.3  & 0.75   &  0.08  &  0.13  &  0.54  & 1    \\
			\hline
			Recall   & 0.37  & 0.75  & 0.12  & 0.25  & 1  & 1  \\
			\hline
			Number of Correct link estimation  & 3  & 6  & 1  & 2  & 6  & 8   \\
			\hline
			\multicolumn{7}{|c|}{Dataset 5}\\
			\hline
		\end{tabular}
		\caption{Performance comparison of PC, GES, LINGAM IC, LINGAM Direct, BIC Exact Search, and SLCD algorithms.}
		\label{tab.3}
	\end{table*}

	\bibliography{mybib}	
\end{document}